\documentclass[twoside]{smgv}
\usepackage{here}
\usepackage{graphicx}
\usepackage{SIunits}

\title{Automatic Pollen Grain and Exine Segmentation from Microscope Images}
\author{Fran\c{c}ois Chung and Tom\'{a}s Rodr\'{i}guez}

\begin{document}
\maketitle

\abstract{In this article, we propose an automatic method for the segmentation of pollen grains from microscope images, followed by the automatic segmentation of their exine. The objective of exine segmentation is to separate the pollen grain in two regions of interest: exine and inner part. A coarse-to-fine approach ensures a smooth and accurate segmentation of both structures. As a rough stage, grain segmentation is performed by a procedure involving clustering and morphological operations, while the exine is approximated by an iterative procedure consisting in consecutive cropping steps of the pollen grain. A snake-based segmentation is performed to refine the segmentation of both structures. Results have shown that our segmentation method is able to deal with different pollen types, as well as with different types of exine and inner part appearance. The proposed segmentation method aims to be generic and has been designed as one of the core steps of an automatic pollen classification framework.}

\section{Introduction}

Bee products are known to have important nutritive and curative properties \cite{Saa-Otero2000}. However, these properties are not currently properly guaranteed in Europe, as there are no standards at European level for certain bee products like pollen and royal jelly. This means that it is possible to find products in the market under these labels without any quality and authenticity control. As a consequence, there has been a significant interest of the scientific community for the study and recognition of bee-related products, such as honey, royal jelly and honeybee pollen.

Pollen is collected by bees in the form of ball-shaped loads known as \emph{pollen loads}. Studies have shown that these loads are \textit{monospecific}, meaning they are composed of grains extracted from the same plant taxon \cite{Corrion2004}. Pollen grains have specific morphological and textural properties that vary from one pollen type to another. These properties referred to as \textit{features} include, among others, the size, shape, color and texture of the grain and are used as discriminant properties for the classification of pollen from different species.

This classification is performed by expert palynologists to study their nutritional and therapeutical properties as well as to characterize their floral and geographical origin. The process of manually separating pollen loads using the mere color information has been and is still widely used by palynologists to classify loads by pollen types \cite{Hodges1984}. However, this process is time-consuming, subjective and requires highly trained palynologists. These issues have been acknowledged by palynologists in the literature \cite{Stillman1996} and many methods to automate the classification process have been proposed in the past.

The main contribution of these methods resides in the use of image processing for the automatic classification of pollen types \cite{DelPozo2012}. For instance, pollen loads have been classified using the only color information extracted from camera images \cite{Chica2012}. However, this macroscopic color-based system is not accurate enough to robustly deal with numerous pollen types, should they feature a similar load color. Nonetheless, this procedure can be applied as a pre-processing step prior to more robust and accurate methods, such as those using microscope images \cite{Wu2008}. So far, the preferred solution for acquiring pollen images is the Light Microscope (LM).

After acquiring the microscope images, the first step in the image processing chain consists in segmenting pollen grains. In practice, this is done by extracting sub-images of pollen grains, from which discriminative features are computed and selected to be representative of the different pollen types. These features are often based on texture and shape, both requiring a different type of segmentation. Texture-based features are usually computed from sub-images of pollen grains separately cropped from the microscope image. Shape-based features must be computed from binary images, or \textit{masks}, separating the pollen grain (foreground) from the rest of the image (background). In both cases, the segmentation is usually performed manually, which is a tedious, time-consuming and subjective task. Pollen grain segmentation methods, whether semi-automatic or automatic \cite{France2000}, have been proposed in the literature, but they do not consider the exine segmentation, thus losing potentially valuable information.

In this work, we propose an automatic method for the segmentation of pollen grains and their exine, both in a coarse-to-fine approach. Segmenting the exine aims at increasing the discriminative power between pollen types by extracting separate features, from the inner part on the one hand and from the exine on the other hand, thus doubling the total number of extracted features. First, sub-images of pollen grains are roughly extracted using a procedure involving clustering and morphological operations (coarse stage). However, our experiments have shown that the exine may not be correctly segmented, depending on the exine properties (e.g. size and color). Furthermore, this rough segmentation is likely to end up with a non-smooth segmentation. This is why the rough segmentation is followed by a segmentation based on snake, or \emph{active contour}, model whose internal forces ensure the segmentation to be smooth (fine stage). Second, the exine is approximated by computing the ratio of edges on the gradient, or \emph{edge}, image of the segmented pollen grain iteratively cropped (coarse stage). Then, the snake algorithm is used again, not only to smooth the segmentation, but also to correctly fit the exine boundary (fine stage).

\section{Data}
\label{sec-data}

\begin{figure}
 \begin{center}
  $\begin{array}{ccccc}
   \multicolumn{1}{l}{} & \multicolumn{1}{l}{} & \multicolumn{1}{l}{} & \multicolumn{1}{l}{} & \multicolumn{1}{l}{}\\
   \includegraphics[scale=0.30]{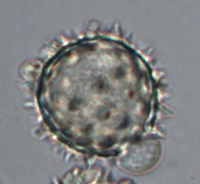} & \includegraphics[scale=0.30]{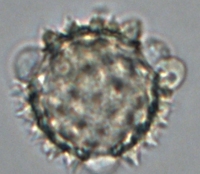} & \includegraphics[scale=0.30]{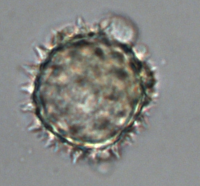} & \includegraphics[scale=0.30]{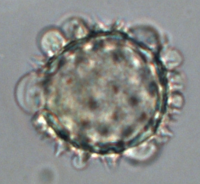} & \includegraphics[scale=0.30]{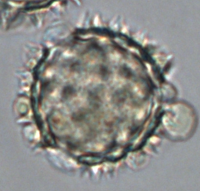}\\
   \includegraphics[scale=0.30]{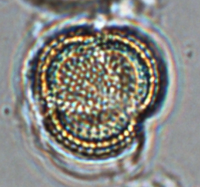} & \includegraphics[scale=0.30]{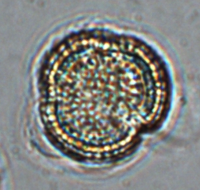} & \includegraphics[scale=0.30]{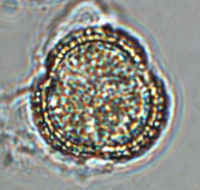} & \includegraphics[scale=0.30]{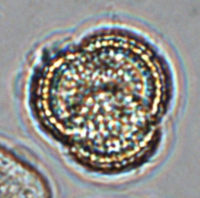} & \includegraphics[scale=0.30]{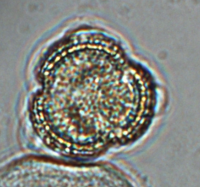}\\
   \includegraphics[scale=0.30]{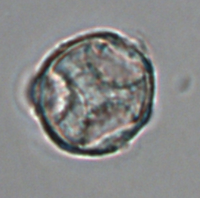} & \includegraphics[scale=0.30]{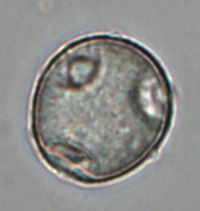} & \includegraphics[scale=0.30]{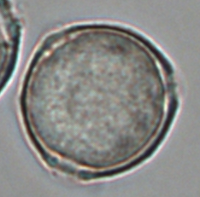} & \includegraphics[scale=0.30]{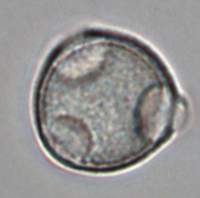} & \includegraphics[scale=0.30]{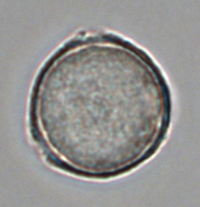}\\
   \includegraphics[scale=0.30]{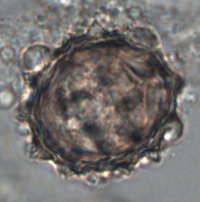} & \includegraphics[scale=0.30]{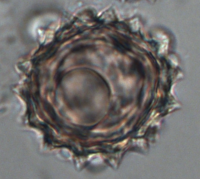} & \includegraphics[scale=0.30]{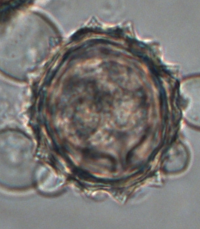} & \includegraphics[scale=0.30]{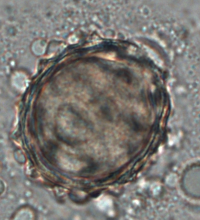} & \includegraphics[scale=0.30]{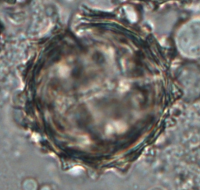}\\
   \includegraphics[scale=0.30]{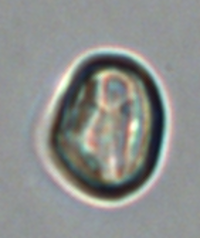} & \includegraphics[scale=0.30]{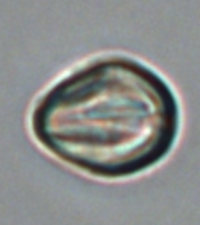} & \includegraphics[scale=0.30]{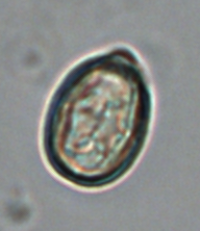} & \includegraphics[scale=0.30]{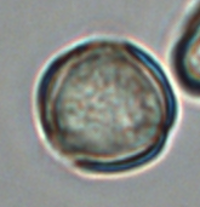} & \includegraphics[scale=0.30]{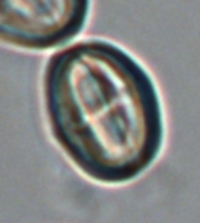}\\
   \includegraphics[scale=0.30]{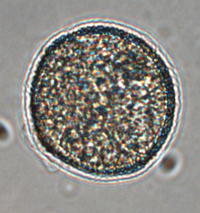} & \includegraphics[scale=0.30]{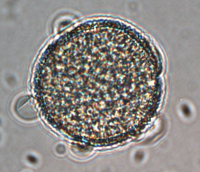} & \includegraphics[scale=0.30]{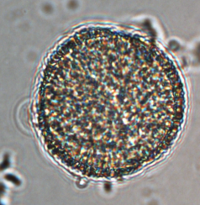} & \includegraphics[scale=0.30]{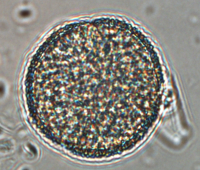} & \includegraphics[scale=0.30]{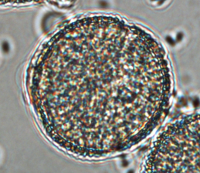}\\
  \end{array}$
 \end{center}
\caption{Microscope sub-images of pollen grains belonging to \emph{Aster}, \emph{Brassica}, \emph{Campanulaceae}, \emph{Carduus}, \emph{Castanea}, and \emph{Cistus} pollen types (from top to bottom, respectively).}
\label{fig-data-1}
\end{figure}

\begin{figure}
 \begin{center}
  $\begin{array}{ccccc}
   \multicolumn{1}{l}{} & \multicolumn{1}{l}{} & \multicolumn{1}{l}{} & \multicolumn{1}{l}{} & \multicolumn{1}{l}{}\\
   \includegraphics[scale=0.30]{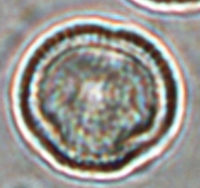} & \includegraphics[scale=0.30]{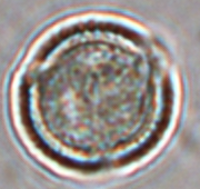} & \includegraphics[scale=0.30]{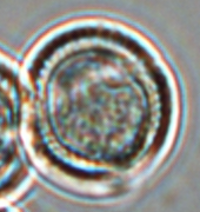} & \includegraphics[scale=0.30]{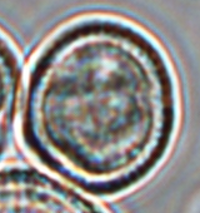} & \includegraphics[scale=0.30]{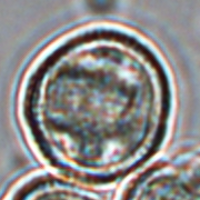}\\
   \includegraphics[scale=0.30]{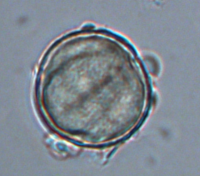} & \includegraphics[scale=0.30]{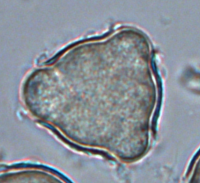} & \includegraphics[scale=0.30]{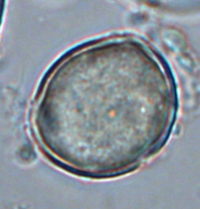} & \includegraphics[scale=0.30]{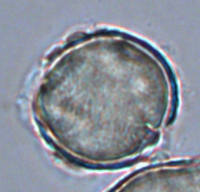} & \includegraphics[scale=0.30]{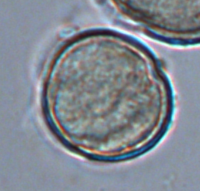}\\
   \includegraphics[scale=0.30]{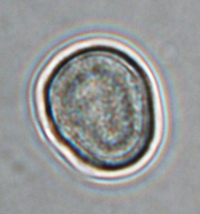} & \includegraphics[scale=0.30]{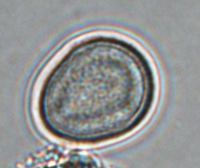} & \includegraphics[scale=0.30]{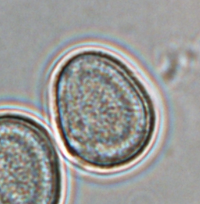} & \includegraphics[scale=0.30]{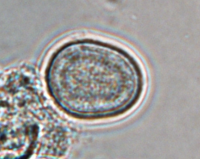} & \includegraphics[scale=0.30]{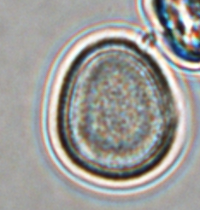}\\
   \includegraphics[scale=0.30]{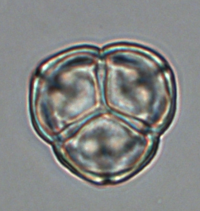} & \includegraphics[scale=0.30]{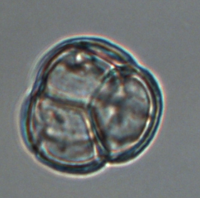} & \includegraphics[scale=0.30]{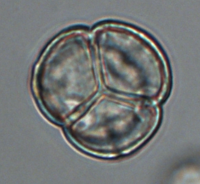} & \includegraphics[scale=0.30]{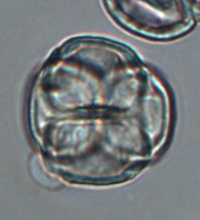} & \includegraphics[scale=0.30]{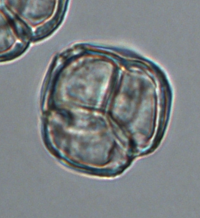}\\
   \includegraphics[scale=0.30]{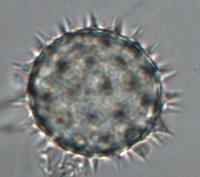} & \includegraphics[scale=0.30]{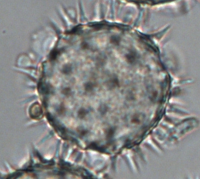} & \includegraphics[scale=0.30]{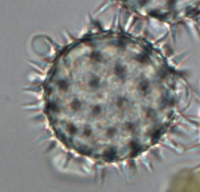} & \includegraphics[scale=0.30]{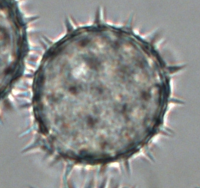} & \includegraphics[scale=0.30]{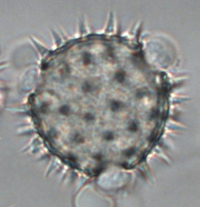}\\
   \includegraphics[scale=0.30]{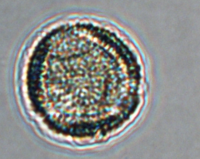} & \includegraphics[scale=0.30]{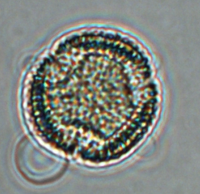} & \includegraphics[scale=0.30]{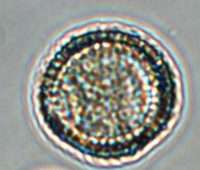} & \includegraphics[scale=0.30]{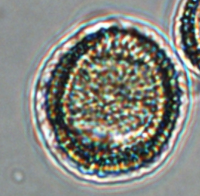} & \includegraphics[scale=0.30]{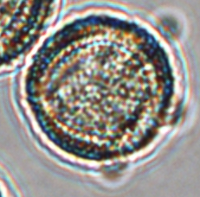}\\
  \end{array}$
 \end{center}
\caption{Microscope sub-images of pollen grains belonging to \emph{Cruciferae}, \emph{Cytisus}, \emph{Echium}, \emph{Ericaceae}, \emph{Helianthus}, and \emph{Olea} pollen types (from top to bottom, respectively).}
\label{fig-data-2}
\end{figure}

\begin{figure}
 \begin{center}
  $\begin{array}{ccccc}
   \multicolumn{1}{l}{} & \multicolumn{1}{l}{} & \multicolumn{1}{l}{} & \multicolumn{1}{l}{} & \multicolumn{1}{l}{}\\
   \includegraphics[scale=0.30]{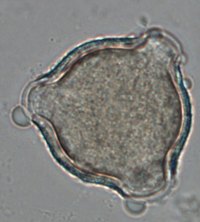} & \includegraphics[scale=0.30]{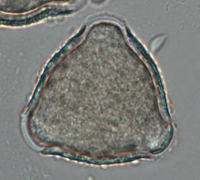} & \includegraphics[scale=0.30]{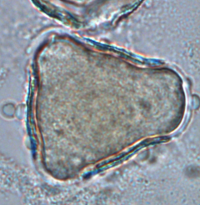} & \includegraphics[scale=0.30]{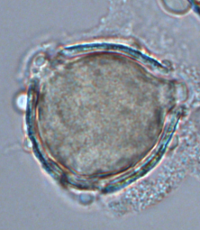} & \includegraphics[scale=0.30]{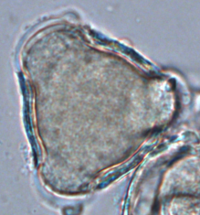}\\
   \includegraphics[scale=0.30]{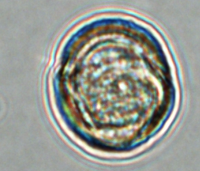} & \includegraphics[scale=0.30]{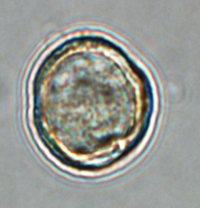} & \includegraphics[scale=0.30]{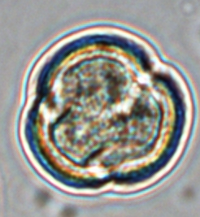} & \includegraphics[scale=0.30]{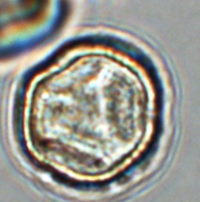} & \includegraphics[scale=0.30]{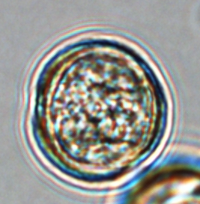}\\
   \includegraphics[scale=0.30]{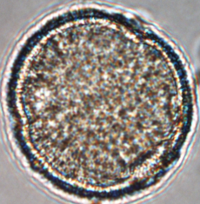} & \includegraphics[scale=0.30]{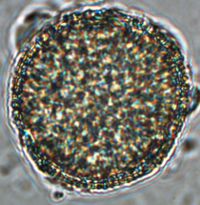} & \includegraphics[scale=0.30]{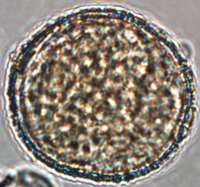} & \includegraphics[scale=0.30]{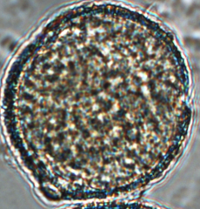} & \includegraphics[scale=0.30]{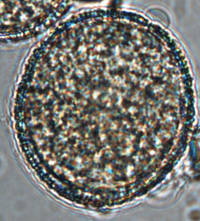}\\
   \includegraphics[scale=0.30]{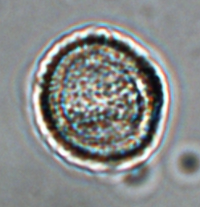} & \includegraphics[scale=0.30]{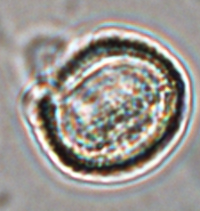} & \includegraphics[scale=0.30]{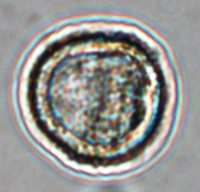} & \includegraphics[scale=0.30]{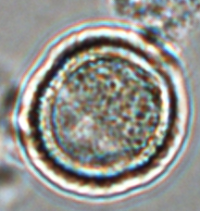} & \includegraphics[scale=0.30]{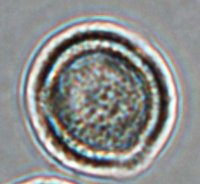}\\
   \includegraphics[scale=0.30]{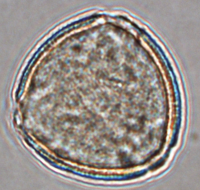} & \includegraphics[scale=0.30]{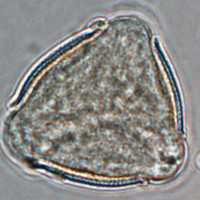} & \includegraphics[scale=0.30]{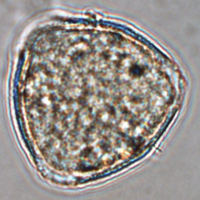} & \includegraphics[scale=0.30]{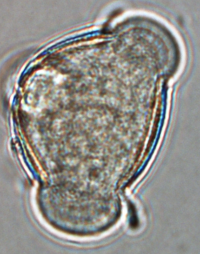} & \includegraphics[scale=0.30]{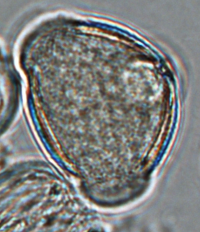}\\
   \includegraphics[scale=0.30]{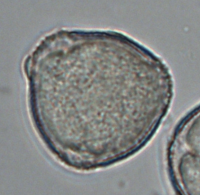} & \includegraphics[scale=0.30]{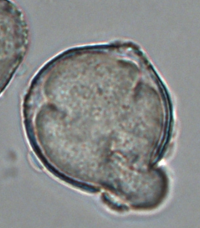} & \includegraphics[scale=0.30]{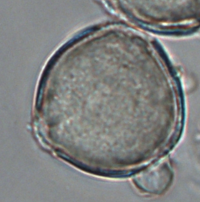} & \includegraphics[scale=0.30]{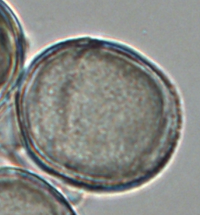} & \includegraphics[scale=0.30]{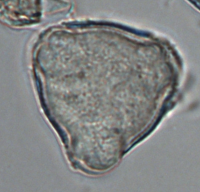}\\
  \end{array}$
 \end{center}
\caption{Microscope sub-images of pollen grains belonging to \emph{Prunus}, \emph{Quercus}, \emph{Reticulo}, \emph{Salix}, \emph{Solanaceae}, and \emph{Teucrium} pollen types (from top to bottom, respectively).}
\label{fig-data-3}
\end{figure}

The number of possible pollen types, or \emph{taxons}, in each country is high. However, there exists a few number of representative types, which make them suitable candidates to test our segmentation method. The following dominant pollen types from Spain, Italy and Turkey, have been selected: \emph{Aster}, \emph{Brassica}, \emph{Campanulaceae}, \emph{Carduus}, \emph{Castanea}, \emph{Cistus}, \emph{Cruciferae}, \emph{Cytisus}, \emph{Echium}, \emph{Ericaceae}, \emph{Helianthus}, \emph{Olea}, \emph{Prunus}, \emph{Quercus}, \emph{Reticulo}, \emph{Salix}, \emph{Solanaceae} and \emph{Teucrium}. In total, we have compiled a pollen image database comprising 18 different pollen types. For each pollen type, an average of 20 microscope images are acquired. The objective is to get enough images to extract a minimum of 100 pollen grain sub-images (one single microscope image may contain up to 30 pollen grains, depending on their size on scattering properties). A brief description of each pollen type is given in Table \ref{tab-data}. Some microscope sub-images of pollen grains are shown in Figure \ref{fig-data-1}, Figure \ref{fig-data-2} and Figure \ref{fig-data-3}.

\begin{table}[t]
\centering
\begin{tabular}{llll}
Type & Size ($\micro\meter$)& Shape                 & Origin\\
\hline
\hline
Aster           & 20-40     & tri-tetra-colporate   & Italy \\
\hline
Brassica        & $\sim25$  & tricolpate            & Spain \\
\hline
Campanulaceae   & 15-30     & tetra-porate          & Spain \\
\hline
Carduus         & 20-50     & tricolporate          & Spain \\
\hline
Castanea        & 8-16      & tricolporate          & Italy/Spain\\
\hline
Cistus          & 26-50     & spheroidal            & Spain\\
\hline
Cruciferae      & 20-40     & tricolpate            & Spain\\
\hline
Cytisus         & 15-30     & tricolporate          & Spain\\
\hline
Echium          & 10-25     & prolate               & Spain\\
\hline
Ericaceae       & 15-25     & tricolporate          & Spain\\
\hline
Helianthus      & 20-40     & tri-colporate         & Bulgary\\
\hline
Olea            & 10-25     & spheroidal            & Spain\\
\hline
Prunus          & 26-50     & spheroidal            & Spain\\
                &           & tricolporate          &       \\
\hline
Quercus         & 26-50     & spheroidal            & Spain\\
\hline
Reticulo        & 45-60     & tricolporate                   & China\\
\hline
Salix           & 16-30     & tricolporate          & Spain\\
                &           & prolate               &      \\
\hline
Solanaceae      & 30-45     & tricolporate                   & China\\
\hline
Teucrium        & 30-40     & tricolporate          & Turkey\\
\hline
\end{tabular}
\caption{Brief description of the pollen types selected to test our segmentation method.}
\label{tab-data}
\end{table}

\section{Material}
\label{sec-material}

To acquire the pollen images, we used the Nikon Eclipse E200-LED bright field microscope featuring \unit{10}{\times}, \unit{20}{\times} and \unit{40}{\times} objectives. The microscope is coupled with the Nikon Digital Sight DS-Fi1 high resolution camera, which acquires the microscope pollen images and transfers them to a computer through a USB connection. This microscope camera is a 5-megapixel charge-coupled device (CCD) capturing color images at 2560$\times$1920 pixel resolution.

Prior to the acquisition of images, pollen grains must be extracted from the pollen loads and placed on a microscope slide. This extraction is performed using ethyl alcohol to clean the slide, silicone grease or glycerine to collect the pollen grains, and a forceps to handle the slides. After the extraction of pollen grains, slides are dried with a heater. The equatorial plane, which is located at the center of the pollen grain \cite{Hesse2009}, is selected as focal plane to acquire the microscope images.

\section{Grain segmentation}
\label{sec-grain-seg}

Pollen grain segmentation consists in extracting sub-images of pollen grains from the microscope image (i.e. one pollen grain per sub-image). In the literature, sub-image extraction is performed through segmentation, whether manual, semi-automatic, or automatic \cite{France2000}. In this work, we propose an automatic segmentation procedure based on a coarse-to-fine approach.

In addition to the sub-image extraction, which consists in cropping the microscope image at pollen grain level (see Figure \ref{fig-grain-seg-fine}, first column), our method extracts the binary image, or \textit{mask}, of the grain out of the sub-image (see Figure \ref{fig-grain-seg-fine}, last column). In the literature, mask extraction is usually not taken into account as the background of microscope images is considered as homogeneous (i.e. featuring a uniform color, or texture), which rather limits its influence on both segmentation and classification results.

\begin{figure}
 \begin{center}
  $\begin{array}{ccccc}
   \multicolumn{1}{l}{} & \multicolumn{1}{l}{} & \multicolumn{1}{l}{} & \multicolumn{1}{l}{} & \multicolumn{1}{l}{}\\
   \includegraphics[scale=0.24]{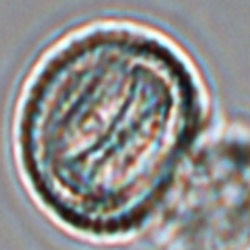} & \includegraphics[scale=0.24]{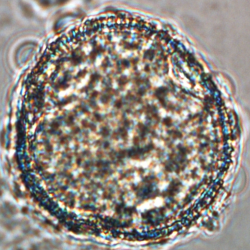} & \includegraphics[scale=0.24]{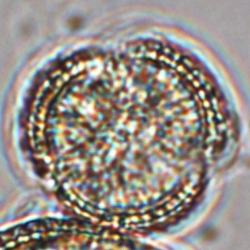} & \includegraphics[scale=0.24]{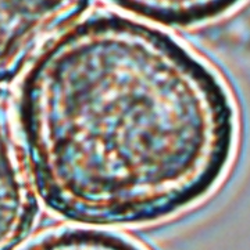} & \includegraphics[scale=0.24]{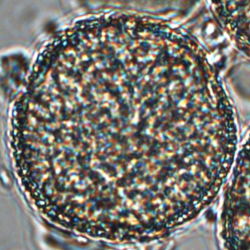}\\
   \mbox{(a)} & \mbox{(b)} & \mbox{(c)} & \mbox{(d)} & \mbox{(e)} \\
  \end{array}$
 \end{center}
\caption{Microscope sub-images of pollen grains whose non-homogeneous background features (a,b) debris, (c,d) grain clusters, and (e) both.}
\label{fig-grain-seg-intro}
\end{figure}

However, the microscopic images may also feature debris that are usually randomly spread in the microscope image. In addition, pollen grains may be close to each other, forming clusters. In practice, this means that, in both cases, the background of the extracted sub-images is likely to be corrupted with non-pollen grain objects, or other pollen grains (see Figure \ref{fig-grain-seg-intro}). This non-homogeneous background must be discarded by means of the mask creation, as it is likely to hamper the quality of the extracted texture features, which may in turn ruin the classification results. In addition, as mentioned in the introduction, mask creation is also a necessary step for the extraction of shape-based features.

\subsection{Coarse stage}

\begin{figure}
 \begin{center}
  $\begin{array}{cccc}
   \multicolumn{1}{l}{} & \multicolumn{1}{l}{} & \multicolumn{1}{l}{} & \multicolumn{1}{l}{}\\
   \includegraphics[scale=0.60]{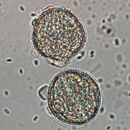} &
   \includegraphics[scale=0.60]{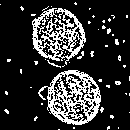} &
   \includegraphics[scale=0.60]{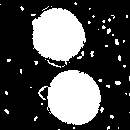} &
   \includegraphics[scale=0.60]{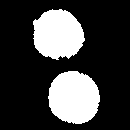}\\
   \mbox{(a)} & \mbox{(b)} & \mbox{(c)} & \mbox{(d)}
  \end{array}$
 \end{center}
\caption{Consecutive steps to automatically segment (a) pollen grains. First, (b) binary classification is used to roughly segment the pollen grains. Then, (c) a hole filling algorithm fills the holes of the inner texture. Finally, (d) opening and closing operations remove the small objects from the image.}
\label{fig-grain-seg-coarse_1}
\end{figure}

First, sub-images of pollen grains are roughly extracted using a procedure involving clustering and morphological operations. As a pre-processing step prior to the segmentation, a contrast-limited adaptive histogram equalization is applied to enhance the contrast of the image. Then, the image is filtered using a median filter to remove noise while preserving edges. Finally, the coarse segmentation of the pollen grains is performed through the following steps (see Figure \ref{fig-grain-seg-coarse_1}).

\begin{enumerate}
  \item Application of a binary classification to the image, so as to roughly separate the pollen grains (foreground) from the rest of the image (background). These two classes are determined by the K-Means algorithm and form the binary image $I_b$.
  \item A hole filling algorithm using 4-connected background neighbours is applied to $I_b$, so as to fill the possible holes featured by the inner texture of the pollen grains.
  \item Opening and closing operations are carried out on $I_b$. The goal is to remove small objects from the image, such as debris, while preserving the shape and size of the pollen grains.
\end{enumerate}

\begin{figure}
 \begin{center}
  $\begin{array}{ccc}
   \multicolumn{1}{l}{} & \multicolumn{1}{l}{}\\
   \includegraphics[scale=0.60]{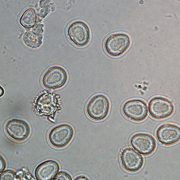} &
   \includegraphics[scale=0.60]{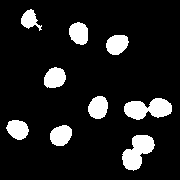} &
   \includegraphics[scale=0.60]{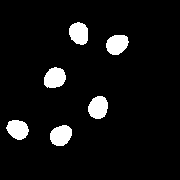}\\
   \mbox{(a)} & \mbox{(b)} & \mbox{(c)}
  \end{array}$
 \end{center}
\caption{After the segmentation of (a) a pollen image, two constraints are applied on the segmented pollen grains: circle-like shape and SD on the image intensity. Results (b) before and (c) after applying these two constraints.}
\label{fig-grain-seg-coarse_2}
\end{figure}

After these three steps, our experiments have shown that small debris are likely to remain in the segmented images. To get rid of these objects, a final post-processing step is necessary. In this work, we propose a post-processing consisting in two constraints applied on pollen grains after segmentation (see Figure \ref{fig-grain-seg-coarse_2}).

First, for pollen grains having a circle-like shape, a constraint is applied on their perimeter (this includes all pollen types from our pollen image database, except \emph{Solanaceae}). The perimeter $P$ of a perfect circle shape is defined by the following formula: $P = 2 \Pi R$, where $R$ is the circle radius. Thus, to ensure that pollen grains have a circle-like shape, their perimeter should be approximated by this formula. Our experiments have shown that constraining the diameter of pollen grains to be smaller than a constant, i.e. $P/2R < 3.55$ ($\approx \Pi$), rejected the majority of debris while keeping pollen grains.

Second, while pollen grains feature textural information in their inner structure, and thus some intensity variation, debris are likely to have a uniform intensity distribution in their inner structure. The second constraint consists thus in ensuring that pollen grains feature a minimal variation in their inner structure using the standard deviation (SD) on the image intensity. Our experiments have shown that constraining this intensity to be greater than a constant, i.e. $\mbox{SD (intensity)} > 20$, gives good results ensuring most of the debris to be discarded.

\subsection{Fine stage}
\label{sec-grain-seg-fine}

After the rough segmentation from the coarse stage, our experiments have shown that most pollen grains are correctly extracted. However, they have also shown that pollen grains may be either slightly over-segmented (i.e. considering the neighboring background as part of the segmentation), or slightly under-segmented (i.e. discarding a part of the exine from the segmentation), depending on the exine properties (e.g. size and color). Furthermore, the coarse stage is likely to end up with a non-smooth segmented grain, as the segmentation is only based on a K-Means clustering of intensity values, i.e. with no consideration for the extracted shape. This is why the rough segmentation is followed by a snake-based segmentation \cite{Xu1997}, with external forces attracting the snake to the exine boundary and internal forces ensuring its contour to be smooth.

\begin{figure}
 \begin{center}
  $\begin{array}{ccccc}
   \multicolumn{1}{l}{} & \multicolumn{1}{l}{} & \multicolumn{1}{l}{} & \multicolumn{1}{l}{}\\
   \includegraphics[scale=0.45]{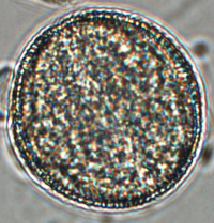} & \includegraphics[scale=0.45]{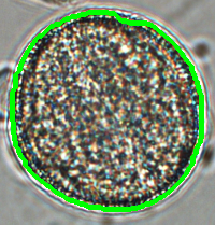} & \includegraphics[scale=0.45]{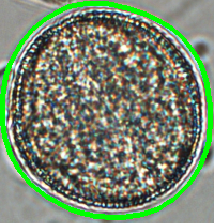} & \includegraphics[scale=0.45]{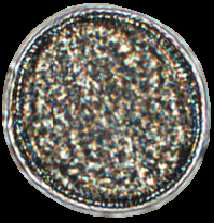}\\[1.5ex]
   \includegraphics[scale=0.45]{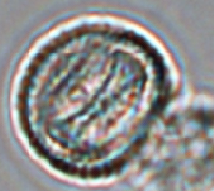} & \includegraphics[scale=0.45]{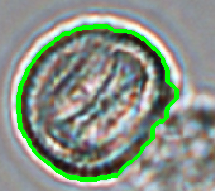} & \includegraphics[scale=0.45]{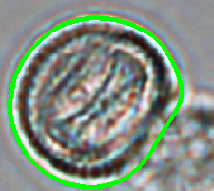} & \includegraphics[scale=0.45]{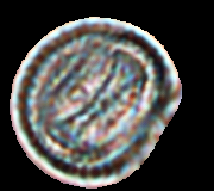}\\[1.5ex]
   \includegraphics[scale=0.45]{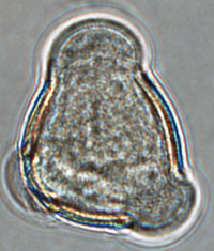} & \includegraphics[scale=0.45]{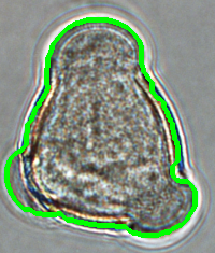} & \includegraphics[scale=0.45]{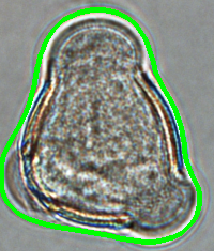} & \includegraphics[scale=0.45]{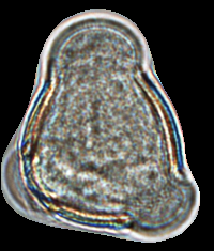}\\[1.5ex]
   \mbox{(a)} & \mbox{(b)} & \mbox{(c)} & \mbox{(d)}\\
  \end{array}$
 \end{center}
\caption{Segmentation of pollen grains belonging to \emph{Cistus} (first row), \emph{Cruciferae} (second row), and \emph{Solanaceae} (third row) pollen types. For each pollen type, (a) original pollen grain, (b) grain segmentation after coarse stage, (c) grain segmentation after fine stage, and (d) output image after retrieving the background (black mask).}
\label{fig-grain-seg-fine}
\end{figure}

The snake is initialized at the perimeter of the mask generated by the rough segmentation. The perimeter is discretized into a set of contour points, which increases the snake's flexibility to fit the exine boundary and speeds up computation time. Empirical tests have shown that keeping one contour point out of 20 consecutive points ensures a good snake initialization. During the segmentation, the snake behaves as a moving contour whose points are attracted to nearby boundaries, such as edges, corners and line terminations. To highlight boundaries, the edge energy image is first extracted by computing the gradient of the original image. Then, a Gradient Vector Flow (GVF), from which the external forces are defined \cite{Xu1997}, is generated from the gradient image. To keep the contour smooth during segmentation, both thin plate energy and balloon force are used as regularization methods. In practice, the snake segmentation consists in 100 iterations, which was found to be sufficient for the snake to fit the grain boundary (see Figure \ref{fig-grain-seg-fine}).

\section{Exine segmentation}
\label{sec-exine-seg}

The separation of the pollen grain in its inner and outer part, known as \emph{the exine}, and composed of one or multiple layers, is a process performed in biology to study both parts separately \cite{Dom�nguez1998,Southworth1988}. In image processing, the pollen grain is considered as a whole, i.e. the region of interest comprises both parts. To classify pollen grains from different types, shape and texture features are usually extracted from the pollen grain, so as to optimize the discriminative power of features from one pollen type to another.

\begin{figure}
 \begin{center}
  $\begin{array}{ccccc}
   \multicolumn{1}{l}{} & \multicolumn{1}{l}{} & \multicolumn{1}{l}{}\\
   \includegraphics[scale=0.40]{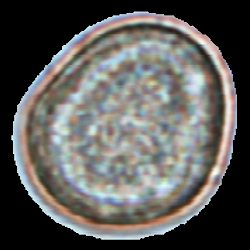} & \includegraphics[scale=0.40]{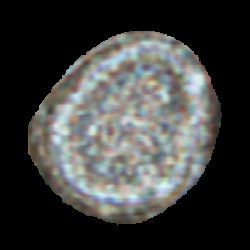} & \includegraphics[scale=0.40]{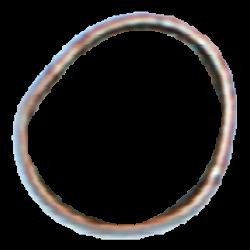}\\[1.5ex]
   \includegraphics[scale=0.40]{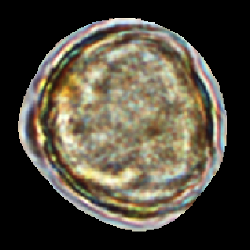} & \includegraphics[scale=0.40]{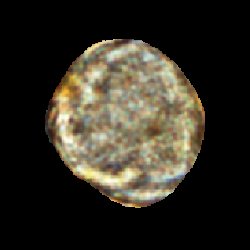} & \includegraphics[scale=0.40]{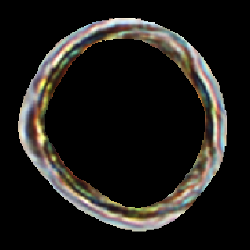}\\[1.5ex]
   \includegraphics[scale=0.40]{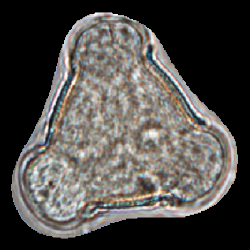} & \includegraphics[scale=0.40]{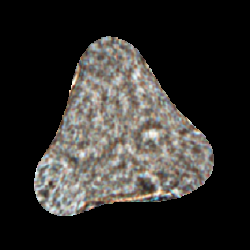} & \includegraphics[scale=0.40]{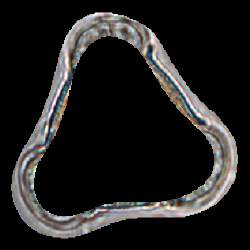}\\
   \mbox{(a)} & \mbox{(b)} & \mbox{(c)}\\
  \end{array}$
 \end{center}
\caption{Difference of texture between the inner part and the exine of pollen grains from: \emph{Echium} (first row), \emph{Retama} (second row), and \emph{Solanaceae} (third row) pollen types. For each pollen type, (a) segmented pollen grain, (b) segmented inner part, and (c) segmented exine.}
\label{fig-exine-seg-intro}
\end{figure}

The objective of exine segmentation is to increase this discriminative power by extracting separate features, from the inner part on the one hand and from the exine on the other hand. Shape features related to the exine, such as the exine thickness, are likely to be discriminant, the exine varying from one pollen type to another. Texture features extracted separately from the inner part and the exine are also likely to be more discriminant, as they will not be extracted from a mixture of texture covering the whole pollen grain, but rather from two separate regions of interest, each of them featuring a more similar texture (see Figure \ref{fig-exine-seg-intro}).

\subsection{Coarse stage}

Unlike the inner part, which usually presents a quite homogeneous texture, the exine may present a rather heterogeneous texture, depending on the pollen type (see Figure \ref{fig-data-1}, Figure \ref{fig-data-2} and Figure \ref{fig-data-3}). This texture heterogeneity is often due to multi-layers constituting the exine, each of them having a different texture. As a consequence, a region growing-based segmentation of the exine is difficult to consider, which is why we propose a gradient-based segmentation method to approximate the exine boundary.

\begin{figure}
 \begin{center}
  $\begin{array}{rl}
   \multicolumn{1}{l}{} & \multicolumn{1}{l}{}\\
   \includegraphics[scale=0.40]{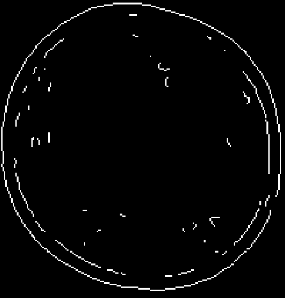} & \includegraphics[scale=0.40]{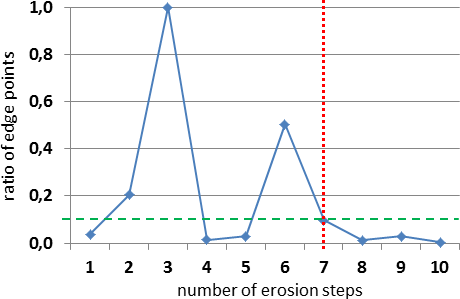}\\
   \includegraphics[scale=0.40]{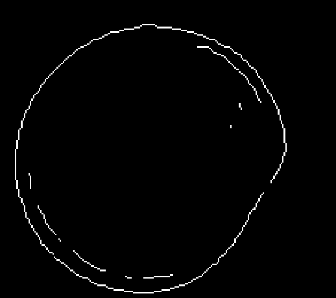} & \includegraphics[scale=0.40]{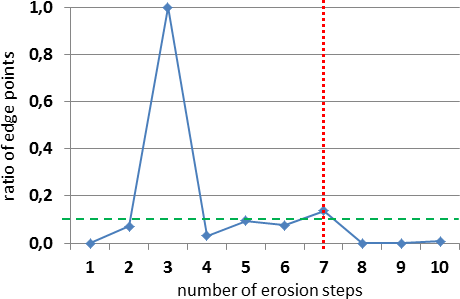}\\
   \includegraphics[scale=0.40]{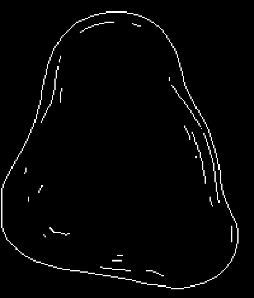} & \includegraphics[scale=0.40]{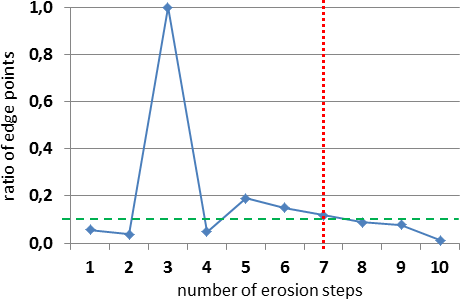}\\
  \end{array}$
 \end{center}
\caption{Coarse stage of the exine segmentation applied on pollen grains belonging to \emph{Cistus} (first row), \emph{Cruciferae} (second row), and \emph{Solanaceae} (third row) pollen types. The edge image is first extracted after anisotropic and Sobel filtering (left column), and then iteratively eroded to compute the ratio of edge points per erosion step (right column). The green horizontal dashed line depicts $\tau_r$, the threshold on the normalized ratio of edge points (here, $\tau_r = 0,1$), from which the boundary between exine and inner part (depicted by the red vertical dotted line) is approximated.}
\label{fig-exine-seg-coarse}
\end{figure}

Our proposed method starts with the extraction of edges from the pollen grain image. First, an anisotropic filtering is applied on the image to smooth the textural information, where small edges might be detected, while preserving the main edge and boundary information. A Sobel filter is applied on the output  image, resulting in the creation of a binary edge image in which the edges (foreground) are extracted from the rest of the image (background). The pollen grain mask extracted in Section \ref{sec-grain-seg} is then applied on the edge image (see Figure \ref{fig-exine-seg-coarse}, left), which is in turn iteratively eroded using a disk-shaped structuring element. At each iteration, the number of edge points in the shrunk image, as well as the total number of points, are calculated. The objective is to compute the ratio of edges points for each iteration.

At the end of this iterative procedure, i.e. when the pollen grain has been shrunk until its complete disappearance, the first derivative is applied on the vector containing the consecutive edge point ratios. The objective is to find a maximum, and its associated number of erosion iterations, corresponding to the approximate exine boundary location, in the hypothesis that the boundary between exine and inner part is featured by a sharp decrease of edge information. In practice, the first derivative is normalized and a threshold on the normalized ratio of edge points $\tau_r$ is defined. Starting from the end of the vector of ratios, and going backwards, the objective is to find a significant gap $\gamma_r$ corresponding to the approximate exine boundary. In practice, the gap is found when the value of the first derivative is higher than $\tau_r$ (see Figure \ref{fig-exine-seg-coarse}, right, in which exine and inner parts are separated at left and right of the red vertical dotted line, respectively). This way, no matter how thick, or variate in terms of texture, is the exine, the method will always consider the first gap as the exine boundary.

\subsection{Fine stage}

\begin{figure}
 \begin{center}
  $\begin{array}{ccccc}
   \multicolumn{1}{l}{} & \multicolumn{1}{l}{} & \multicolumn{1}{l}{} & \multicolumn{1}{l}{}\\
   \includegraphics[scale=0.45]{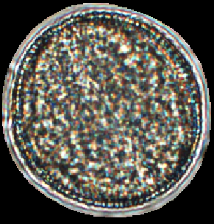} & \includegraphics[scale=0.45]{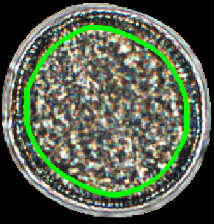} & \includegraphics[scale=0.45]{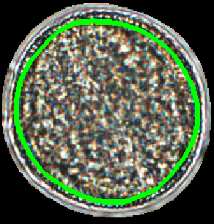} & \includegraphics[scale=0.45]{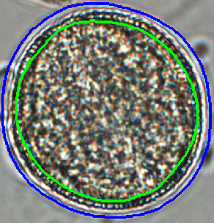}\\[1.5ex]
   \includegraphics[scale=0.45]{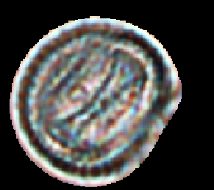} & \includegraphics[scale=0.45]{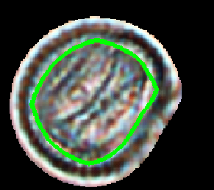} & \includegraphics[scale=0.45]{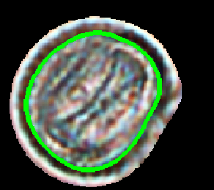} & \includegraphics[scale=0.45]{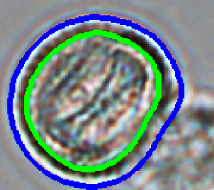}\\[1.5ex]
   \includegraphics[scale=0.45]{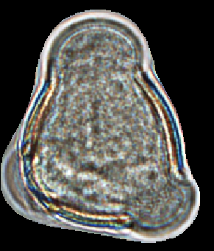} & \includegraphics[scale=0.45]{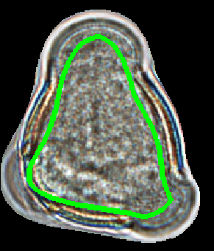} & \includegraphics[scale=0.45]{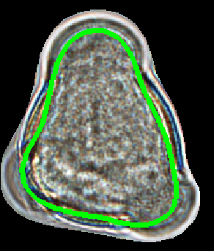} & \includegraphics[scale=0.45]{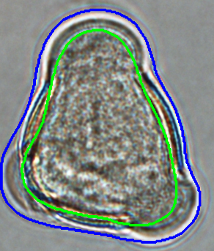}\\[1.5ex]
   \mbox{(a)} & \mbox{(b)} & \mbox{(c)} & \mbox{(d)}\\
  \end{array}$
 \end{center}
\caption{Exine segmentation of pollen grains belonging to \emph{Cistus} (first row), \emph{Cruciferae} (second row), and \emph{Solanaceae} (third row) pollen types. For each pollen type, (a) segmented pollen grain, (b) exine segmentation after coarse stage, (c) exine segmentation after fine stage, and (d) final segmentation depicting both grain (blue contour) and exine (green contour) segmentations.}
\label{fig-exine-seg-fine}
\end{figure}

At the end of the coarse stage, the boundary between exine and inner part is approximated by iteratively eroding the segmented pollen grain with the disk-shaped structuring element from the coarse stage and the number of erosion iterations associated to the gap $\gamma_r$. The fine stage consists in the same snake-based segmentation method presented in Section \ref{sec-grain-seg-fine}. As for the pollen grain segmentation, the snake is initialized using the discretized perimeter of the mask generated by the rough exine segmentation, then attracted to the exine boundary using the same internal and external forces, and the same number of iterations (see Figure \ref{fig-exine-seg-fine}).

\section{Results}

\begin{figure}
 \begin{center}
  $\begin{array}{ccccc}
   \multicolumn{1}{l}{} & \multicolumn{1}{l}{} & \multicolumn{1}{l}{} & \multicolumn{1}{l}{} & \multicolumn{1}{l}{}\\
   \includegraphics[scale=0.30]{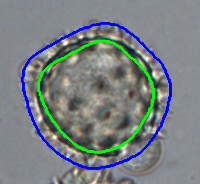} & \includegraphics[scale=0.30]{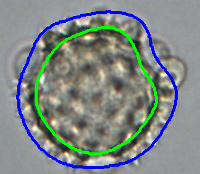} & \includegraphics[scale=0.30]{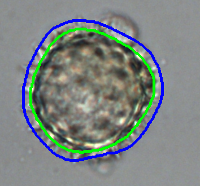} & \includegraphics[scale=0.30]{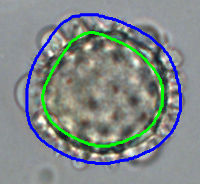} & \includegraphics[scale=0.30]{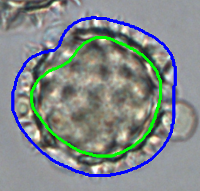}\\
   \includegraphics[scale=0.30]{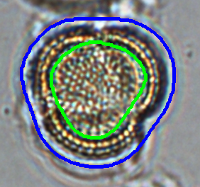} & \includegraphics[scale=0.30]{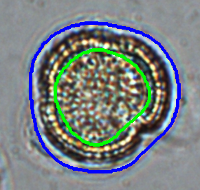} & \includegraphics[scale=0.30]{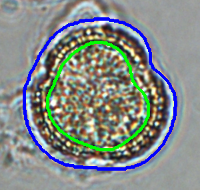} & \includegraphics[scale=0.30]{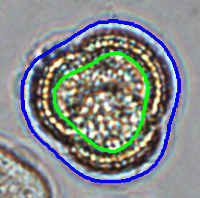} & \includegraphics[scale=0.30]{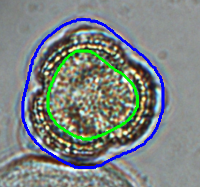}\\
   \includegraphics[scale=0.30]{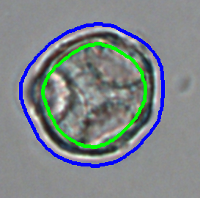} & \includegraphics[scale=0.30]{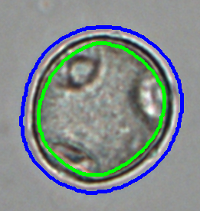} & \includegraphics[scale=0.30]{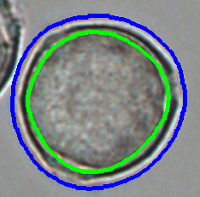} & \includegraphics[scale=0.30]{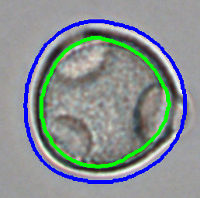} & \includegraphics[scale=0.30]{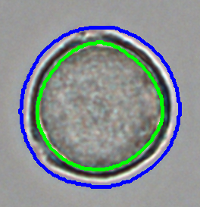}\\
   \includegraphics[scale=0.30]{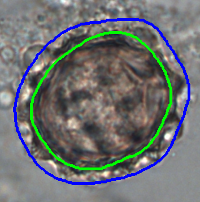} & \includegraphics[scale=0.30]{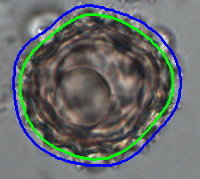} & \includegraphics[scale=0.30]{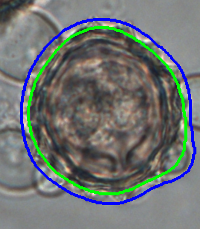} & \includegraphics[scale=0.30]{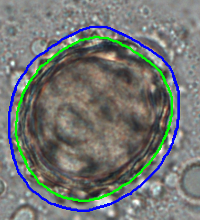} & \includegraphics[scale=0.30]{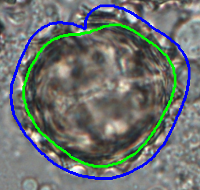}\\
   \includegraphics[scale=0.30]{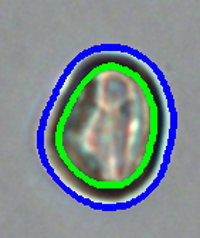} & \includegraphics[scale=0.30]{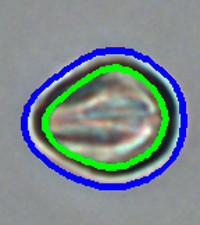} & \includegraphics[scale=0.30]{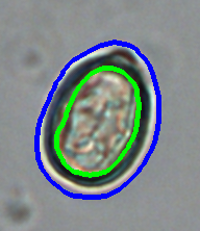} & \includegraphics[scale=0.30]{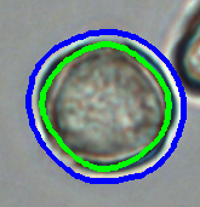} & \includegraphics[scale=0.30]{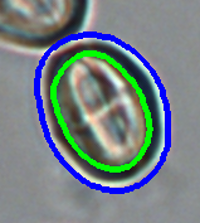}\\
   \includegraphics[scale=0.30]{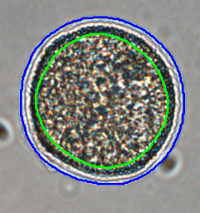} & \includegraphics[scale=0.30]{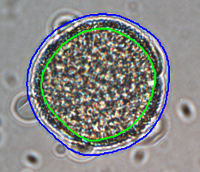} & \includegraphics[scale=0.30]{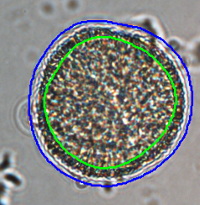} & \includegraphics[scale=0.30]{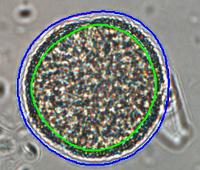} & \includegraphics[scale=0.30]{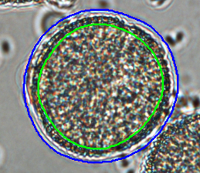}\\
  \end{array}$
 \end{center}
\caption{Final segmentation of pollen grains (blue contour) and their exine (green contour). The original sub-images of the pollen grains are depicted in Figure \ref{fig-data-1}. Pollen types are \emph{Aster}, \emph{Brassica}, \emph{Campanulaceae}, \emph{Carduus}, \emph{Castanea}, and \emph{Cistus} (from top to bottom, respectively).}
\label{fig-results-1}
\end{figure}

\begin{figure}
 \begin{center}
  $\begin{array}{ccccc}
   \multicolumn{1}{l}{} & \multicolumn{1}{l}{} & \multicolumn{1}{l}{} & \multicolumn{1}{l}{} & \multicolumn{1}{l}{}\\
   \includegraphics[scale=0.30]{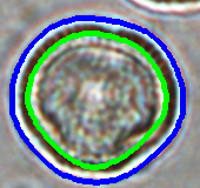} & \includegraphics[scale=0.30]{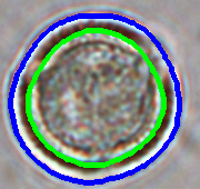} & \includegraphics[scale=0.30]{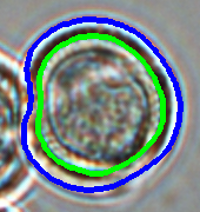} & \includegraphics[scale=0.30]{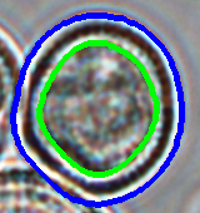} & \includegraphics[scale=0.30]{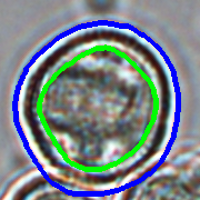}\\
   \includegraphics[scale=0.30]{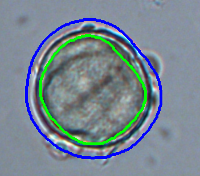} & \includegraphics[scale=0.30]{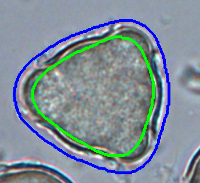} & \includegraphics[scale=0.30]{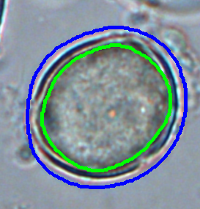} & \includegraphics[scale=0.30]{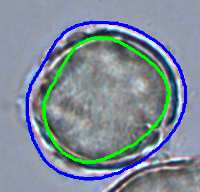} & \includegraphics[scale=0.30]{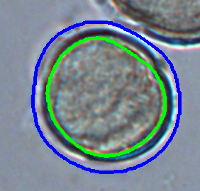}\\
   \includegraphics[scale=0.30]{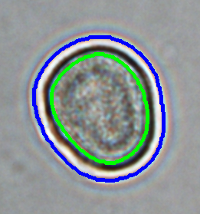} & \includegraphics[scale=0.30]{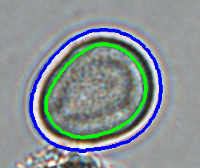} & \includegraphics[scale=0.30]{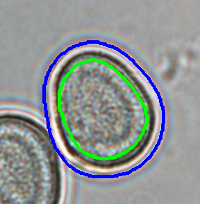} & \includegraphics[scale=0.30]{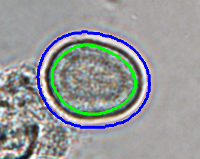} & \includegraphics[scale=0.30]{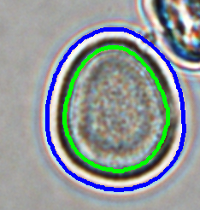}\\
   \includegraphics[scale=0.30]{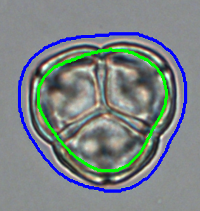} & \includegraphics[scale=0.30]{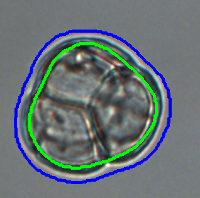} & \includegraphics[scale=0.30]{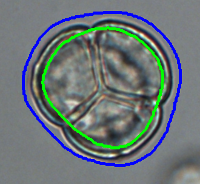} & \includegraphics[scale=0.30]{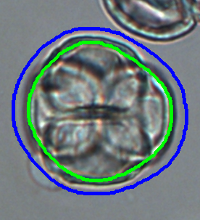} & \includegraphics[scale=0.30]{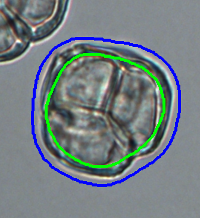}\\
   \includegraphics[scale=0.30]{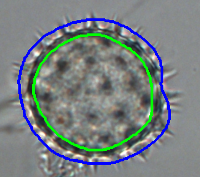} & \includegraphics[scale=0.30]{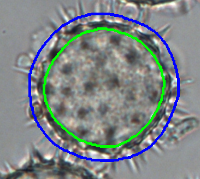} & \includegraphics[scale=0.30]{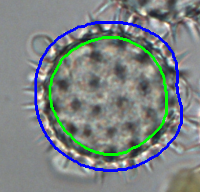} & \includegraphics[scale=0.30]{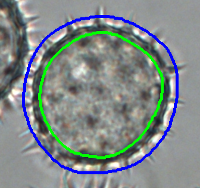} & \includegraphics[scale=0.30]{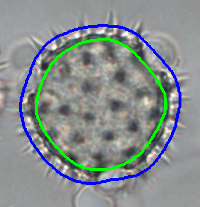}\\
   \includegraphics[scale=0.30]{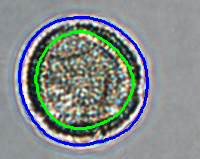} & \includegraphics[scale=0.30]{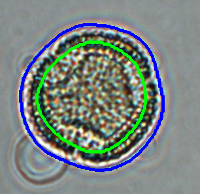} & \includegraphics[scale=0.30]{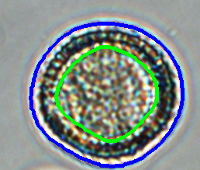} & \includegraphics[scale=0.30]{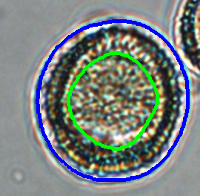} & \includegraphics[scale=0.30]{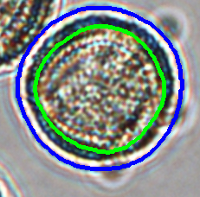}\\
  \end{array}$
 \end{center}
\caption{Final segmentation of pollen grains (blue contour) and their exine (green contour). The original sub-images of the pollen grains are depicted in Figure \ref{fig-data-2}. Pollen types are \emph{Cruciferae}, \emph{Cytisus}, \emph{Echium}, \emph{Ericaceae}, \emph{Helianthus}, and \emph{Olea} (from top to bottom, respectively).}
\label{fig-results-2}
\end{figure}

\begin{figure}
 \begin{center}
  $\begin{array}{ccccc}
   \multicolumn{1}{l}{} & \multicolumn{1}{l}{} & \multicolumn{1}{l}{} & \multicolumn{1}{l}{} & \multicolumn{1}{l}{}\\
   \includegraphics[scale=0.30]{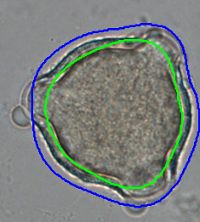} & \includegraphics[scale=0.30]{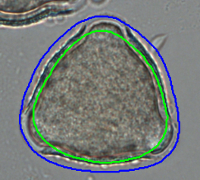} & \includegraphics[scale=0.30]{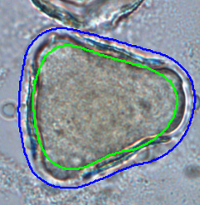} & \includegraphics[scale=0.30]{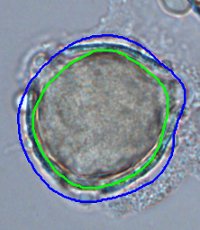} & \includegraphics[scale=0.30]{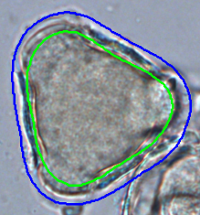}\\
   \includegraphics[scale=0.30]{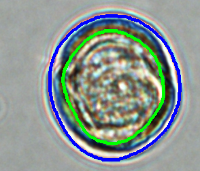} & \includegraphics[scale=0.30]{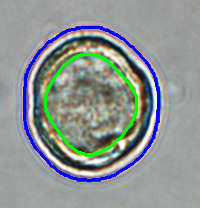} & \includegraphics[scale=0.30]{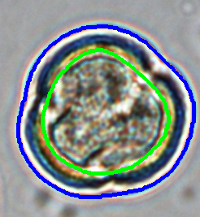} & \includegraphics[scale=0.30]{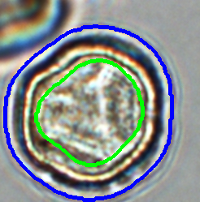} & \includegraphics[scale=0.30]{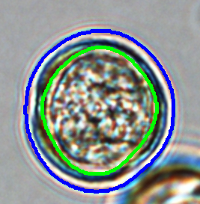}\\
   \includegraphics[scale=0.30]{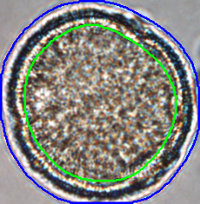} & \includegraphics[scale=0.30]{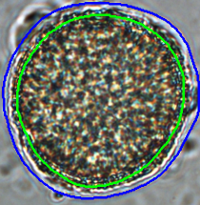} & \includegraphics[scale=0.30]{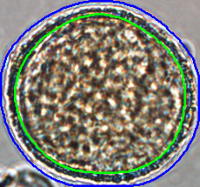} & \includegraphics[scale=0.30]{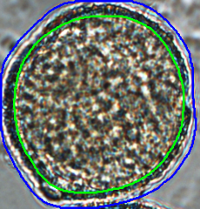} & \includegraphics[scale=0.30]{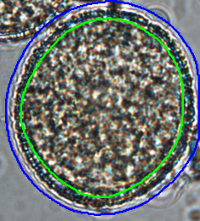}\\
   \includegraphics[scale=0.30]{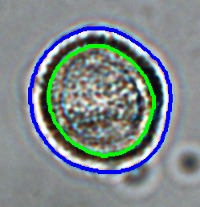} & \includegraphics[scale=0.30]{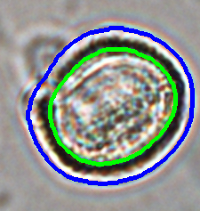} & \includegraphics[scale=0.30]{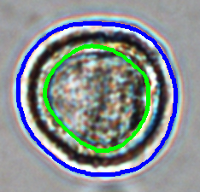} & \includegraphics[scale=0.30]{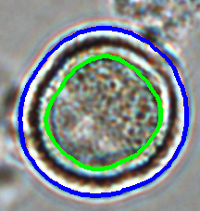} & \includegraphics[scale=0.30]{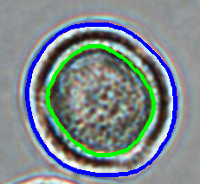}\\
   \includegraphics[scale=0.30]{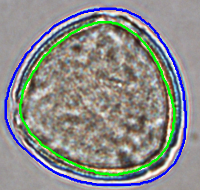} & \includegraphics[scale=0.30]{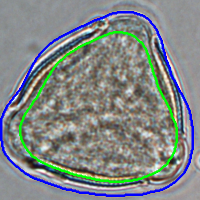} & \includegraphics[scale=0.30]{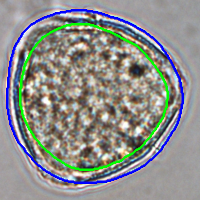} & \includegraphics[scale=0.30]{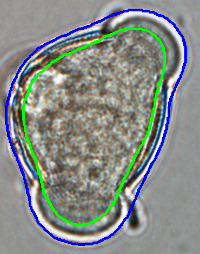} & \includegraphics[scale=0.30]{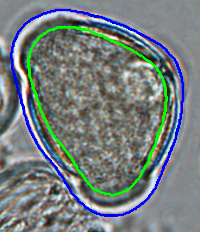}\\
   \includegraphics[scale=0.30]{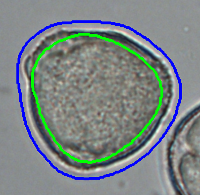} & \includegraphics[scale=0.30]{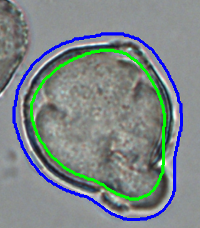} & \includegraphics[scale=0.30]{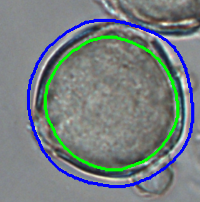} & \includegraphics[scale=0.30]{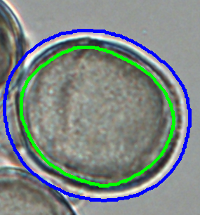} & \includegraphics[scale=0.30]{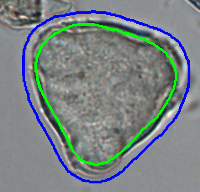}\\
  \end{array}$
 \end{center}
\caption{Final segmentation of pollen grains (blue contour) and their exine (green contour). The original sub-images of the pollen grains are depicted in Figure \ref{fig-data-3}. Pollen types are \emph{Prunus}, \emph{Quercus}, \emph{Reticulo}, \emph{Salix}, \emph{Solanaceae}, and \emph{Teucrium} (from top to bottom, respectively).}
\label{fig-results-3}
\end{figure}

We have applied our automatic method for the segmentation of both pollen grains and their exine on the image database presented in Section \ref{sec-data}. Pollen images were acquired with the material described in Section \ref{sec-material}, and both grain and exine segmentations were performed by the methods explained in Section \ref{sec-grain-seg} and Section \ref{sec-exine-seg}, respectively. The final segmentation of some pollen grains and their exine is depicted in Figure \ref{fig-results-1}, Figure \ref{fig-results-2} and Figure \ref{fig-results-3}.

The majority of pollen grains from the image database have been correctly segmented (see Table \ref{tab-results}). Regarding the exine segmentation, our method has shown to be robust enough to deal with different types of exine. The majority of pollen types feature a double-layered exine composed of a black inner wall and a white outer wall, i.e. \emph{Campanulaceae} (Figure \ref{fig-results-1}, 3rd row), \emph{Castanea} (Figure \ref{fig-results-1}, 5th row), \emph{Cruciferae} (Figure \ref{fig-results-2}, 1st row), \emph{Echium} (Figure \ref{fig-results-2}, 3rd row), \emph{Quercus} (Figure \ref{fig-results-3}, 2nd row) and \emph{Salix} (Figure \ref{fig-results-3}, 4th row). \emph{Cruciferae}, \emph{Echium} and \emph{Salix} have exine walls quite thick, while \emph{Campanulaceae}, \emph{Castanea} and \emph{Quercus} have an inner wall thicker than the outer wall. In both cases, the segmentation method performs as expected, taking into account the two walls of the exine.

\begin{table}[t]
\centering
\begin{tabular}{cccccc}
Aster & Brassica & Campanulaceae & Carduus & Castanea & Cistus\\
\hline
0.82 & 0.84 & 0.87 & 0.88 & 0.94 & 0.88\\
\hline
\hline
Cruciferae & Cytisus & Echium & Ericaceae & Helianthus & Olea\\
\hline
0.68 & 0.94 & 0.96 & 0.86 & 0.66 & 0.86\\
\hline
\hline
Prunus & Quercus & Reticulo & Salix & Solanaceae & Teucrium \\
\hline
0.77 & 0.85 & 0.59 & 0.79 & 0.64 & 0.87\\
\end{tabular}
\caption{Segmentation accuracy for the pollen types of our pollen image database, which is calculated as the number of correctly segmented pollen grains (i.e. both grains and their exine) with respect to the total count of pollen grains present in the microscope images.}
\label{tab-results}
\end{table}

Another common type of exine consists in a wall whose texture is similar, but denser, than the inner part, such as for \emph{Brassica} (Figure \ref{fig-results-1}, 2nd row), \emph{Cistus} (Figure \ref{fig-results-1}, 6th row), \emph{Olea} (Figure \ref{fig-results-2}, 6th row) and \emph{Reticulo} (Figure \ref{fig-results-3}, 3rd row). Note that these pollen types also feature a thin white outer wall. Regarding \emph{Aster} (Figure \ref{fig-results-1}, 1st row), \emph{Carduus} (Figure \ref{fig-results-1}, 4th row) and \emph{Helianthus} (Figure \ref{fig-results-2}, 5th row), their exine is made of spike-like structures and the segmentation of their pollen grains takes into account the surrounding region containing most of these structures (note that, for \emph{Helianthus}, there is a slight under-segmentation as some spike peaks are not taken into account). Other segmentation challenges were encountered. The exine of \emph{Cytisus} (Figure \ref{fig-results-2}, 2nd row), \emph{Prunus} (Figure \ref{fig-results-3}, 1st row), \emph{Solanaceae} (Figure \ref{fig-results-3}, 5th row), \emph{Teucrium} (Figure \ref{fig-results-3}, 6th row) may present exine discontinuity. Regarding \emph{Ericaceae} (Figure \ref{fig-results-2}, 4th row), not only is the exine thin, but it also features a texture very similar to the inner part. Our experiments have shown that the segmentation is able to deal with these challenges.

\begin{figure}
 \begin{center}
  $\begin{array}{ccccc}
   \multicolumn{1}{l}{} & \multicolumn{1}{l}{} & \multicolumn{1}{l}{} & \multicolumn{1}{l}{} & \multicolumn{1}{l}{}\\
   \includegraphics[scale=0.30]{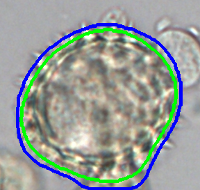} & \includegraphics[scale=0.30]{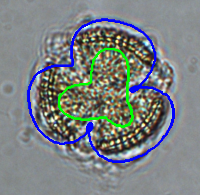} & \includegraphics[scale=0.30]{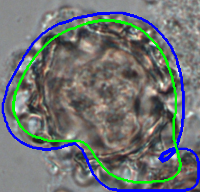} & \includegraphics[scale=0.30]{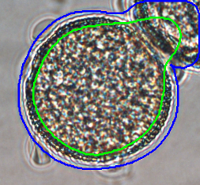} & \includegraphics[scale=0.30]{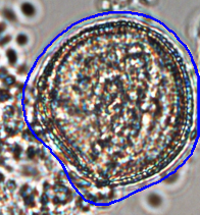}\\[1.5ex]
   \mbox{(a)} & \mbox{(b)} & \mbox{(c)} & \mbox{(d)} & \mbox{(e)}\\
  \end{array}$
 \end{center}
\caption{Some segmentation errors for both pollen grains (blue contour) and their exine (green contour): under-segmentation of (a) the exine and (b) both grain and exine due to the lack of salient edges, over-segmentation due to the nearby (c) dust or (d) pollen grain, and (e) absence of exine segmentation due to the lack of salient boundary between exine and inner part.}
\label{fig-bad-results}
\end{figure}

However, segmentation results are closely related to the quality of the morphological features exhibited by pollen grains, depending on which working hypotheses may or may not be met. For instance, should the boundary between exine and inner part be unclear due to the lack if salient edges, the exine is likely to be under-segmented and close to the grain boundary, where edge information is usually more important (see Figure \ref{fig-bad-results}a). The worst case scenario is when there is no edge information around the grain boundary, in which case only the pollen grain can be segmented (see Figure \ref{fig-bad-results}e). In case there is no complete edge information at the grain boundary, the pollen grain \emph{per se} is likely to be under-segmented (see Figure \ref{fig-bad-results}b). In this case, the under-segmentation occurs at the coarse stage, and the snake is thus initialized at the wrong location. As a consequence, both grain and exine are under-segmented. Conversely, over-segmentation may occur in the presence of nearby dust (see Figure \ref{fig-bad-results}c) or pollen grain (see Figure \ref{fig-bad-results}d), should they be close enough to each other, feature a similar texture and have no salient boundary edges to distinguish one from the other. It should be noted though, that the exine segmentation seems to be invariant to the different types of inner part appearance, such as smooth (e.g. \emph{Prunus} and \emph{Teucrium}), textured (e.g. \emph{Brassica} and \emph{Cistus}), pattern-like (e.g. \emph{Aster} and \emph{Helianthus}), and different depending on the pollen polarity (e.g. \emph{Campanulaceae} and \emph{Castanea}).

\section{Conclusion}

To our knowledge, this work is the first to present an exine segmentation method using image processing techniques. We have proposed an automatic method for the segmentation of pollen grains from microscope images, followed by the automatic segmentation of their exine. The objective of exine segmentation is to separate the pollen grain in two regions of interest: exine and inner part. A coarse-to-fine approach ensures a smooth and accurate segmentation of both structures. The method starts with the rough extraction of pollen grains from the microscope image by means of a procedure involving clustering and morphological operations (coarse stage). Then, a snake-based segmentation is performed to ensure a more accurate and smoother segmentation (fine stage). In a second step, the exine is approximated by using the gradient image and an iterative procedure consisting in consecutive cropping steps of the pollen grain (coarse stage). Finally, the snake algorithm is once again used to refine the exine segmentation (fine stage).

Our results have shown that our segmentation method is able to deal with the different pollen types from the image database, as well as with the different types of exine appearance. Some of them present segmentation challenges, such as a double-layered exine (made of an inner wall of different texture and thickness compared to the outer wall) and a exine made of spike-like structures. In the latter case, although the segmentation may be slightly under-segmented at spike peak level (see \emph{Helianthus}, Figure \ref{fig-results-2}, 5th row), the segmented exine can be used to approximate the exine thickness and its texture. In most cases, the method shows a good performance in dealing with these segmentation challenges and performs as expected for both grain and exine segmentation, i.e. the segmentation has shown to be invariant to the different types of exine and inner part appearance.

Overall, the proposed segmentation method aims to be generic and used in a complete pollen classification framework. The first objective is to ease the segmentation process \emph{per se}. Instead of manually segmenting both pollen grains and their exine, which is a tedious, time-consuming and subjective task, the proposed segmentation method extracts them in an automatic fashion. The second objective is to extract both pollen grains and their exine in a complete pollen classification framework in which discriminant features are then computed to classify the grains by pollen type. So far in the literature, discriminant features have been computed from the complete pollen grain (exine and inner part). By separating exine and inner part, two different sets of features may be extracted to reinforce the discriminant power of features, so as to improve the global segmentation accuracy.

\paragraph{Acknowledgements}The present work has been developed under the framework of the APIFRESH project (FP7-SME-2008-2/243594), which is co-funded by the European Commission. The authors would like to thank Rafael Redondo, Amelia Gonzalez and Cristina Pardo for their help in the microscope image acquisition.

\bibliographystyle{plain}
\bibliography{biblio}

\begin{thebibliography}{10}

\bibitem{Corrion2004}
P.~Carri\'{o}n, E.~Cernadas, J.F. G\'{a}lvez, M.~Dami\'{a}n, and
  P.~S\'{a}-Otero.
\newblock Classification of honeybee pollen using a multiscale texture
  filtering scheme.
\newblock {\em Machine Vision and Applications}, 15(4):186--193, 2004.

\bibitem{Chica2012}
M.~Chica and P.~Campoy.
\newblock Discernment of bee pollen loads using computer vision and one-class
  classification techniques.
\newblock {\em Journal of Food Engineering}, 112(1–2):50--59, 2012.

\bibitem{DelPozo2012}
M.~del Pozo-Ba{\~n}os, J.R. Ticay-Rivas, J.~Cabrera-Falc{\'o}n, J.~Arroyo, C.M.
  Travieso-Gonz{\'a}lez, L.~S{\'a}nchez-Chavez, S.T. P{\'e}rez, J.B. Alonso,
  and M.~Ram{\'\i}rez-Bogantes.
\newblock {\em Image Processing for Pollen Classification}, chapter~19, pages
  493--508.
\newblock Intech, 2012.

\bibitem{France2000}
I.~France, A.W.G. Duller, G.A.T. Duller, and H.F. Lamb.
\newblock A new approach to automated pollen analysis.
\newblock {\em Quaternary Science Reviews}, 19(6):537--546, 2000.

\bibitem{Hesse2009}
M.~Hesse, H.~Halbritter, M.~Weber, R.~Buchner, and R.Zetter.
\newblock {\em Pollen Terminology: An Illustrated Handbook}.
\newblock Springer Vienna, 2009.

\bibitem{Hodges1984}
D.~Hodges.
\newblock {\em The pollen loads of the honeybee: a guide to their
  identification by colour and form}.
\newblock International Bee Research Association, 1984.

\bibitem{Saa-Otero2000}
M.P. Saa-Otero, E.~D\'{i}az-Losada, and E.~Fern\'{a}ndez-G\'{o}mez.
\newblock Analysis of fatty acids, proteins and ethereal extract in honeybee
  pollen - considerations of their floral origin.
\newblock {\em Grana}, 39(4):175--181, 2000.

\bibitem{Southworth1988}
D.~Southworth.
\newblock Isolation of exines from gymnosperm pollen.
\newblock {\em American Journal of Botany}, 75(1):15--21, 1988.

\bibitem{Stillman1996}
E.C. Stillman and J.R. Flenley.
\newblock The needs and prospects for automation in palynology.
\newblock {\em Quaternary Science Reviews}, 15(1):1--5, 1996.

\bibitem{Wu2008}
Q.~Wu, F.~Merchant, and K.R. Castleman.
\newblock {\em Microscope Image Processing}.
\newblock Elsevier/Academic Press, 2008.

\bibitem{Xu1997}
Chenyang Xu and J.L. Prince.
\newblock Gradient vector flow: a new external force for snakes.
\newblock In {\em Proceedings of the IEEE Computer Society Conference on
  Computer Vision and Pattern Recognition, 1997}, pages 66--71, 1997.

\end{thebibliography}
\end{document}